\documentclass[letterpaper]{article}
\usepackage{aaai}
\usepackage{times}
\usepackage{helvet}
\usepackage{courier}
\usepackage{graphicx}
\usepackage{algorithm}      
\usepackage{algorithmicx}
\usepackage{algpseudocode}
\usepackage{booktabs} 
\usepackage{amssymb}
\usepackage{bm}  
\usepackage{enumitem}       
\usepackage{amsmath}        
\usepackage{algpseudocode}
\usepackage{multirow}
\usepackage{booktabs}
\usepackage{wasysym}
\usepackage{natbib}  
\begin{document}
%
\title{Conditional Video Generation for Effective Video Compression}
\author{Fangqiu Yi, Jingyu Xu, Jiawei Shao, Chi Zhang, Xuelong Li*.\\
\{yifq1, xujy70, shaojw2, zhangc120\}@chinatelecom.cn, xuelong\_li@ieee.org\\
Institute of Artificial Intelligence (TeleAI), China Telecom\\
\\
}
\maketitle
\begin{abstract}

Traditional and neural video compression methods often struggle at ultra-low bitrates, yielding video content with poor perceptual quality characterized by blurring, blocking, and color shifts. This limitation stems from their focus on pixel-level accuracy rather than extracting critical information essential for human visual perception. To overcome this limitation, we propose a novel video compression framework that reframes video compression as a conditional generation task, where a generative model synthesizes video from sparse yet informative signals.  Our approach introduces three key modules: (1) Multi-granular conditioning that captures both static scene structure and dynamic spatio-temporal cues; (2) Compact representations designed for efficient transmission without sacrificing semantic richness; (3) Multi-condition training with modality dropout and role-aware embeddings, which prevent over-reliance on any single modality and enhance robustness. Extensive experiments show that our method significantly outperforms both traditional and neural codecs on perceptual quality metrics such as Fréchet Video Distance (FVD) and LPIPS, especially under high compression ratios.

\end{abstract}

\section{Introduction}

The exponential growth of video content across streaming platforms, social networks, teleconferencing, and augmented reality applications has created unprecedented demand for effective compression techniques. Current video compression standards, including H.266/VVC \citep{zhang2020fast} and AV1~\citep{chen2020overview}, have achieved substantial improvements through decades of engineering refinement, employing hybrid coding strategies that combine motion estimation, transform coding, and entropy modeling. However, these approaches rely on largely handcrafted components within rigid codec architectures, limiting their adaptability to diverse application requirements.

Most traditional and neural compression pipelines operate under a fundamental assumption: the pursuit of pixel-level fidelity to ensure reconstructed frames match the original input as closely as possible. While this approach suits applications requiring exact reproduction—such as scientific imaging or professional video editing—we argue that strict fidelity is often unnecessary for perceptual consumption scenarios. In applications like user-generated content, entertainment streaming, or virtual conferencing, perceptual consistency—visual coherence aligned with human perception—matters more than exact pixel reconstruction. Relaxing pixel-perfect accuracy requirements creates opportunities for aggressive compression while enabling new trade-offs between bitrate and perceptual quality.
Data-driven approaches have emerged as promising alternatives to traditional codecs. Recent advances in neural image and video compression leverage encoder-decoder architectures and learned entropy models to achieve competitive rate-distortion performance. However, most methods remain constrained by deterministic reconstruction requirements and often exhibit suboptimal perceptual quality, particularly at low bitrates, manifesting as blurring, blocking artifacts, and color degradation.

Concurrently, generative models—especially diffusion models—have demonstrated state-of-the-art performance in image and video synthesis. This paradigm shifts focus from encoding pixel-level residuals to achieving content-faithful reconstruction under strict bitrate constraints by leveraging the strong priors of generative models and compact spatio-temporal guidance. While existing methods~\citep{zhang2025videocodingmeetsmultimodal, wan2024m3cvccontrollablevideocompression, wu2023sketch} model spatio-temporal information via textual prompts, keyframes, or basic visual cues, we contend these representations are insufficient for high-fidelity video reconstruction. This limitation confines their applicability to narrow domains—such as  human-face video~\citep{chen},  human-body video~\citep{wang2022, wang2023}, or small motion video~\citep{yin2024} — where scene complexity remains constrained.  


We explore this question through a diffusion-based compression framework that introduces three core innovations. First, we employ multi-granular signals that capture both static scene structure such as auto selected keyframes and semantic descriptions, and dynamic information including human motion, optical flow, and panoptic segmentation. Second, we design compact, transmission-efficient representations for these signals that serve as minimal yet perceptually informative inputs to the decoder. Third, we develop a multi-condition training strategy that incorporates signal dropout and role-aware embeddings, enabling the model to remain robust even when certain signals are unavailable or degraded.

We evaluate our method on standard video benchmarks using perceptual metrics including Fréchet Video Distance (FVD) and LPIPS. Results demonstrate substantial improvements over both traditional and learning-based codecs in perceptual quality, particularly at high compression ratios. These findings suggest that conditional generative models offer a promising paradigm for perception-centric video compression, where semantic compactness and visual plausibility supersede strict pixel accuracy.

To summarize, our main contributions are threefold:
\begin{itemize}
    \item First, we design an efficient and modular compression framework, incorporating multi-granular conditioning and compact, transmission-friendly representations to guide reconstruction with minimal overhead.
    \item Second, we deploy a multi-condition training strategy enabling the model to remain robust even when certain signals are unavailable or degraded.
    \item Third, we demonstrate state-of-the-art perceptual performance on standard benchmarks, significantly outperforming traditional and neural codecs under high compression ratios, as measured by FVD and LPIPS.
\end{itemize}

\begin{figure*}[t]
    \centering
    \includegraphics[width=0.95\textwidth]{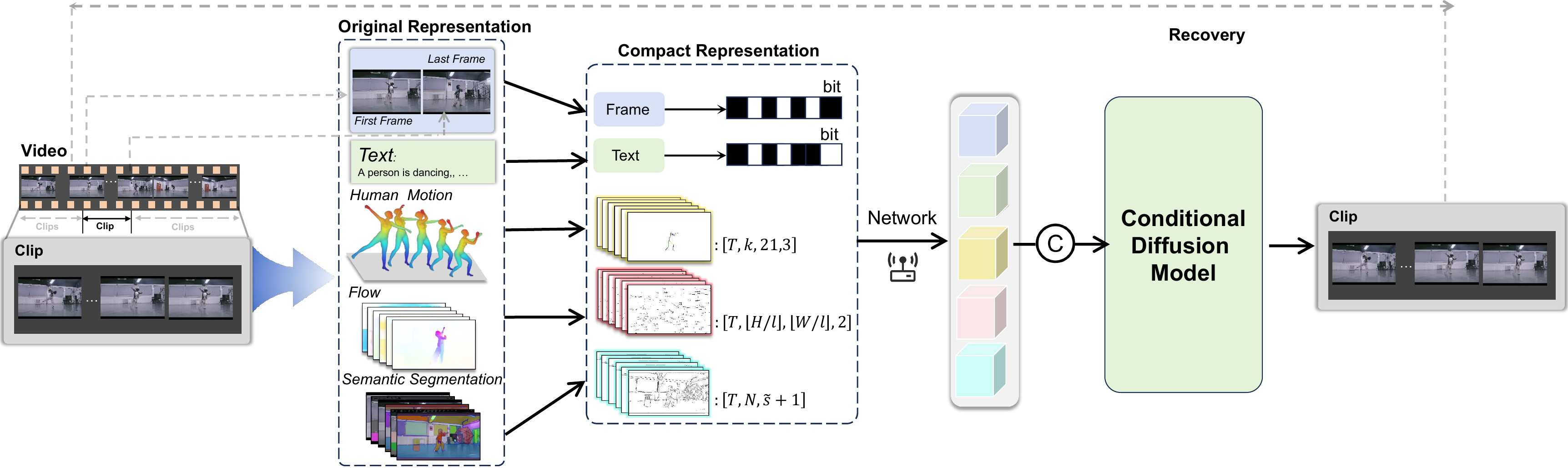}
    \caption{Our framework processes input videos through three sequential stages: First, a keyframe selection module partitions the video into consecutive clips. Second, clip-specific conditions are compressed—the first frame, last frame, and textual descriptions via entropy coding while segmentation sequences, human motion data, and optical flow are converted to compact representations. Finally, at the decoder, a conditional diffusion model reconstructs each clip using all decompressed conditions to generate the output video.}
    \label{fig:pipeline}
\end{figure*}

\section{Related Work}
\subsection{Video Compression}
Video compression aims to reduce redundant information in video sequences while preserving critical visual content, enabling efficient storage and transmission across applications such as streaming services, video conferencing, and surveillance. 
With the success of deep learning in image compression \citep{mishra2022deep}, neural video compression has emerged as a prominent research area, leveraging neural networks to optimize rate-distortion trade-offs.
Residual coding-based approaches \citep{choi2019deep} generate predicted frames from previously decoded ones and encode the residual between predicted and current frames.
However, their reliance on simple subtraction for inter-frame redundancy reduction leads to suboptimal performance. 
Lu et al. \citep{li2021deep} pioneered this direction by replacing traditional codec modules with neural networks in an end-to-end framework. 
The DCVC series \citep{li2021deep} exemplifies this paradigm, with DCVC-DC \citep{li2023neural} and DCVC-FM \citep{li2024neural} outperforming traditional codecs like ECM \citep{karadimitriou1996set}. 
While these neural video compression methods have made significant strides in improving rate-distortion performance as measured by pixel-level metrics, they often overlook the perceptual quality of the reconstructed videos, which is crucial for human viewing experience. 
This gap between pixel-level metrics and perceptual quality led Blau et al. \citep{blau2019rethinking} to highlight a ``rate-distortion-perception" trade-off, spurring research into perceptual video compression. 

\subsection{Conditional Video Generation}
Conditional video generation aims to synthesize videos adhering precisely to external conditions (appearance, layout, motion) while maintaining spatiotemporal consistency. 
Early methods extended image diffusion models with text guidance, enabling creative generation but lacking fine-grained detail and motion control.
Subsequent approaches incorporated stronger conditions: Image animation methods used initial frames but often produced static results.
Methods using low-level dense signals (e.g., depth or edge sequences: Gen-1 \citep{esser2023structure}, ControlVideo \citep{zhao2025controlvideo}, VideoComposer \citep{wang2023videocomposer}) improved control but proved impractical. 
Object trajectory or layout control emerged via strokes (DragNUWA \citep{yin2023dragnuwa}), coordinates (MotionCtrl \citep{wang2024motionctrl}), or bounding boxes. 
Training-based trajectory methods (TrackGo \citep{zhou2025trackgo}) were costly with limited gains, while training-free attention manipulation (FreeTraj \citep{qiu2024freetraj}) suffered inaccuracies. 
Critically, existing methods remain fragmented: each control modality typically requires specialized inputs or architectural changes, highlighting the need for unified, adaptable frameworks.

\subsection{Diffusion Models for Video Compression }
Diffusion-based compression has advanced rapidly in the image domain, laying groundwork for video applications.
Wu et al. \citep{wu2023sketch} transmitted sketches and text descriptions to guide diffusion-based reconstruction, while Careil et al. \citep{careil2023towards} used vector-quantized latents and captions for decoding. 
Relic et al. \citep{relic2025bridging} optimized efficiency by framing quantization noise removal as a denoising task with adaptive steps.
However, extending these advances to video faces key challenges: integrating foundational diffusion models into existing video coding paradigms (e.g., conditional coding) without disrupting efficiency, mitigating slow inference, and enabling multi-bitrate support for varying latent distortion levels.
To address these, this work focuses on diffusion-based video compression, emphasizing dynamic and static condition controls to balance compression ratio and perceptual quality. By harmonizing these conditions, the proposed method enhances reconstruction fidelity across bitrates while preserving coding efficiency—bridging the gap between diffusion-based image compression success and unmet needs in video coding.

\section{Method}

In this section, we formally introduce our proposed video compression framework, which consists of three key stages, as illustrated in Figure~\ref{fig:pipeline}. Each stage is designed to extract, compress, and reconstruct video content in a perceptually consistent and bandwidth-efficient manner.
\vspace{-4mm}
\paragraph{Keyframe Selection and Clip Segmentation.} The input video is first processed by a keyframe selection module, which partitions it into sequential, non-overlapping clips. For each clip, the first and last frames are selected as keyframes, serving as spatial anchors for subsequent reconstruction. These keyframes also act as natural segmentation points, enabling each clip to be independently encoded and decoded.
\vspace{-4mm}
\paragraph{Conditional Feature Extraction and Compression.} For the intermediate (non-key) frames within each clip, we extract a set of multi-granular conditional features that capture complementary aspects of the visual content. These include semantic captions (textual descriptions of the scene), panoptic segmentation maps (spatial structure), human motion sequences, and optical flow (temporal dynamics). The extracted features are then compressed into a compact representation suitable for transmission, preserving the essential cues needed for high-quality reconstruction while minimizing bitrate.
\vspace{-2mm}
\paragraph{Conditional Frame Generation at the Decoder.} Each clip is transmitted as a combination of compressed keyframes and compact conditional representations. At the decoder, we employ a pre-trained multi-conditional diffusion model to reconstruct the intermediate frames by jointly conditioning on the received keyframes and auxiliary signals. This generative process ensures both spatial fidelity and temporal coherence across the reconstructed video sequence.

We now provide detailed descriptions of each component in the following subsections.






\subsection{1. Keyframe Selection and Clip Segmentation}
\label{sec:keyframe_selection}

Our keyframe selection strategy employs dual-criterion detection to balance compression efficiency and reconstruction quality. Formally, given video sequence $\mathcal{V} = \{f_1, f_2, \dots, f_T\}$ with $T$ frames, we identify keyframes through:

\begin{enumerate}[label=(\roman*)]
    \item \textbf{Shot boundary detection:} Frame $f_i$ is selected when identified as a shot transition frame using TransNetV2 \citep{soucek2020transnetv2}. We compute shot transition probability $p_i = \mathcal{T}(f_i)$ where $\mathcal{T}$ denotes the pretrained TransNetV2 model, and designate $f_i$ as a keyframe when $p_i > 0.5$.
    \vspace{-2mm}
    \item \textbf{Fixed-interval sampling:} Frame $f_i$ is selected if $i - i_{\text{prev}} \geq w$, where $w$ is a tunable hyperparameter and $i_{\text{prev}}$ denotes the previous keyframe index. Smaller $w$ values increase keyframe density (improving reconstruction quality at higher bitrates) while larger $w$ reduces transmission overhead.
    
\end{enumerate}


\subsection{2. Conditional Feature Extraction}
Empirical studies have established that, spatiotemporal control signals significantly enhance motion dynamics, spatial alignment, and temporal consistency - critical factors for perceptual fidelity in video reconstruction. Besides the textual descriptions obtained through VLLM, we select three complementary conditional representations whose compact forms balance reconstruction quality against transmission bandwidth.

\begin{algorithm}[ht]
\caption{Keyframe-based Clip Segmentation}
\label{alg:clip_segmentation}
\begin{algorithmic}[1]
\Require Video frames $\mathcal{V} = \{f_1, f_2, \dots, f_T\}$,
\Statex \hspace*{4mm} Hyperparameter $w$,
\Statex \hspace*{4mm} Shot detector $\mathcal{T}$
\Ensure Clips $\mathcal{C} = \{ [start_m,  end_m] \}_{m=1}^M$

\State $\mathcal{K} \gets \{0\}$  \Comment{Initialize with first frame}
\State $\mathit{last\_key} \gets 0$
\State $\mathit{prev\_type} \gets \texttt{null}$

\For{$i \gets 1$ \textbf{to} $T$}
    \If{$(i - \mathit{last\_key}) \geq w$ \textbf{or} $\mathcal{T}(f_i) > 0.5$}
        \State $\mathcal{K} \gets \mathcal{K} \cup \{i\}$
        \If{$(i - \mathit{last\_key}) \geq w$}
            \State $\mathit{prev\_type} \gets \texttt{interval}$
        \Else
            \State $\mathit{prev\_type} \gets \texttt{shot}$
        \EndIf
        \State $\mathit{last\_key} \gets i$
    \EndIf
\EndFor
\State $\mathcal{K} \gets \mathcal{K} \cup \{T\}$  \Comment{Add last frame}

\State $\mathcal{C} \gets \emptyset$
\State $\mathit{keys} \gets \text{SORT}(\mathcal{K})$ \Comment{Sorted keyframe indices}
\For{$j \gets 1$ \textbf{to} $|\mathit{keys}| - 1$}
    \State $s \gets \mathit{keys}[j-1]$
    \State $e \gets \mathit{keys}[j]$
    \If{$\mathit{prev\_type} = \texttt{shot}$}
        \State $s \gets s + 1$  \Comment{Offset for shot boundary}
    \EndIf
    \State $\mathcal{C} \gets \mathcal{C} \cup \{ [s, e] \}$
    \State $\mathit{prev\_type} \gets \text{type of } \mathit{keys}[j]$
\EndFor
\end{algorithmic}
\end{algorithm}

\subsection{Segmentation Sequences}
Segmentation sequences provide crucial geometric scaffolding by preserving object boundaries and spatial relationships across frames. This explicit structural prior prevents shape distortion during generation and maintains consistent object interactions throughout temporal transitions. 
We extract per-frame panoptic segmentation map using Mask2Former \citep{cheng2021mask2former}. To achieve high compression, we first extract its external contour using border tracing, and then Approximate each contour with $n$-th order Bézier curves:

$$
B(t) = \sum_{i=0}^{n} \binom{n}{i} (1-t)^{n-i} t^i P_i, \quad t \in [0,1]
$$
where $P_i$ are optimized control points. Finally, we retain only the $N$ longest contours per frame. Fig.~\ref{fig:demo}(a) shows the original segmentation map and the fitting results obtained with different values of $N$, where 8th-order Bézier curve is applied. This compact representation preserves essential object topology while largely reducing storage since only the Bézier parameters need to be saved.


\subsection{Human Motion Representation}
Human motion sequences are essential for human-centeric videos~\citep{zhang2024vast} since it significantly reduces artifacts in articulated movements and maintains natural temporal coherence during complex actions like walking or dancing. We represent human motion using 3D SMPL sequences~\citep{smpl}. Then, we select 21 joints that represents human kinetics and project those 3D joints to 2D image-coordinates. By calculating the proportion of the area occupied by 2D joints in the image coordinates, we set different thresholds to filter out small human poses. Fig.~\ref{fig:demo}(b) shows the human motion results obtained with different area-threshold ${\xi}$. The 2D joints coordinates are transmitted over the network.

\subsection{Optical Flow Representation}
Optical flow fields explicitly encode dense displacement vectors between pixels, providing critical motion guidance. We compute optical flow using RAFT~\citep{raft}. However, the cost of transmitting pixel-level optical flow is prohibitively expensive. We believe that the representation of optical flow should be based on blocks of pixels rather than individual pixels. Specifically, we define sampling stride $l$ (hyperparameter controlling compression ratio) and then sample flow vectors at regular intervals.
\[
    \mathcal{G} = \left\{ \mathcal{G}(x,y) \mid x = \lfloor i \cdot l \rfloor,  y = \lfloor j \cdot l \rfloor \right\}
\]
where $i = 1, 2, \dots, \lfloor H/l \rfloor$, $j = 1, 2, \dots, \lfloor W/l \rfloor$, $\mathcal{G}$ is the extracted optical flow map. 

Under large sampling strides, bilinear interpolation produces significant errors in recovered optical flow fields. Therefore, we employ a flow-arrow visualization approach where arrow direction indicates motion orientation at sampled points and arrow length represents flow magnitude. Fig.~\ref{fig:demo}(c) demonstrates this flow visualization across different sampling strides.

\begin{figure*}[ht]
    \centering
    \includegraphics[width=0.75\textwidth]{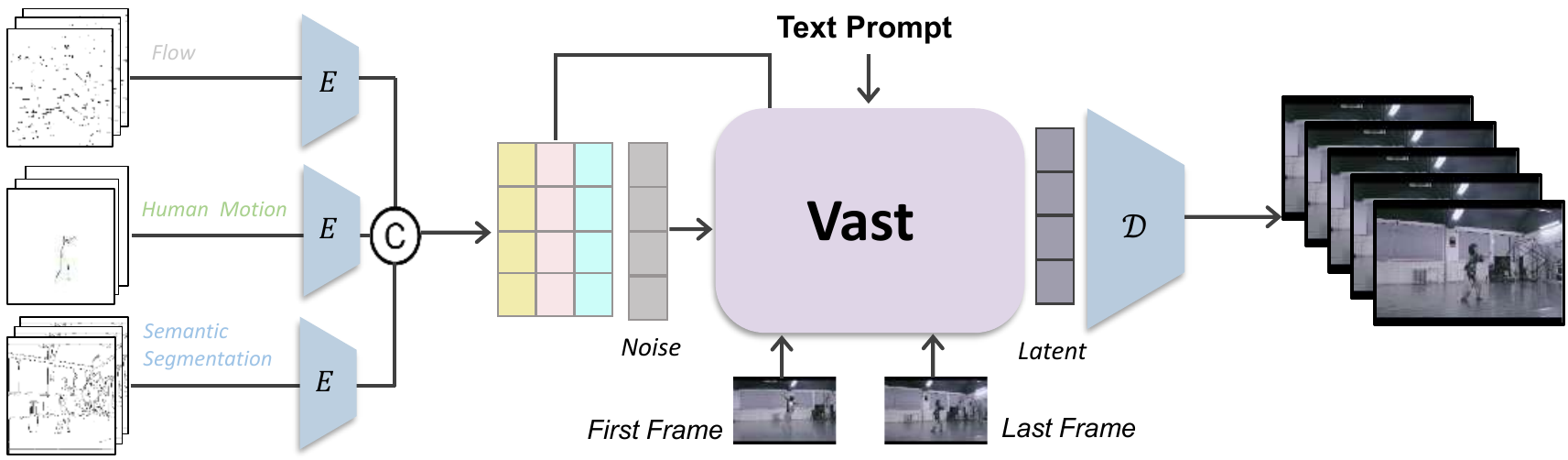}
    \vspace{-1mm}
    \caption{Our diffusion model converts optical flow, segmentation, and motion into visual modalities, encodes them via VAE, and concatenates the latent codes with noise as input to the diffusion backbone.}
    \label{fig:diff}
\end{figure*}

\subsection{Compression Calculation}
For a clip, the first and last frames are encoded into a bitstream using a state-of-the-art image compression method, such as LIC~\citep{li2025on}, along with text, with a size set to $Q$KB. The representations of the remaining three conditions are directly encoded in numeric form, with each number represented using bfloat16 (2 Bytes). The compressed bitrate (KBps) of our framework is calculated per video clip as:

\begin{equation}
\begin{split}
R = \frac{Q \cdot \text{fps}}{T} 
&+ \frac{2 \cdot \text{fps}}{1024} \cdot \Bigg[ \\
&\quad \phi(\xi) \cdot 21 \cdot 2 + 2 \left\lfloor\frac{H}{l}\right\rfloor \left\lfloor\frac{W}{l}\right\rfloor + 2N(n+1) \Bigg] 
\end{split}
\end{equation}

\begin{itemize}[leftmargin=*,noitemsep]
    \item $Q$: Size of compressed first frame and last frame + text (KB)
    \item $T$: Clip length
    \item $\text{fps}$: Frame rate (frames/sec) of the clip
    \item $k$: The number of people pose remaining under threshold $\phi({\xi})$
    \item $H,W$: Frame dimensions (pixels)
    \item $l$: Flow sampling stride
    \item $N$: The number of Bézier curves
    \item $n$: Order of Bézier curve (default: 8)
\end{itemize}

\subsection{3. Conditional Frame Generation at Decoder}
\begin{figure*}[t]
    \centering
    \includegraphics[width=0.85\textwidth]{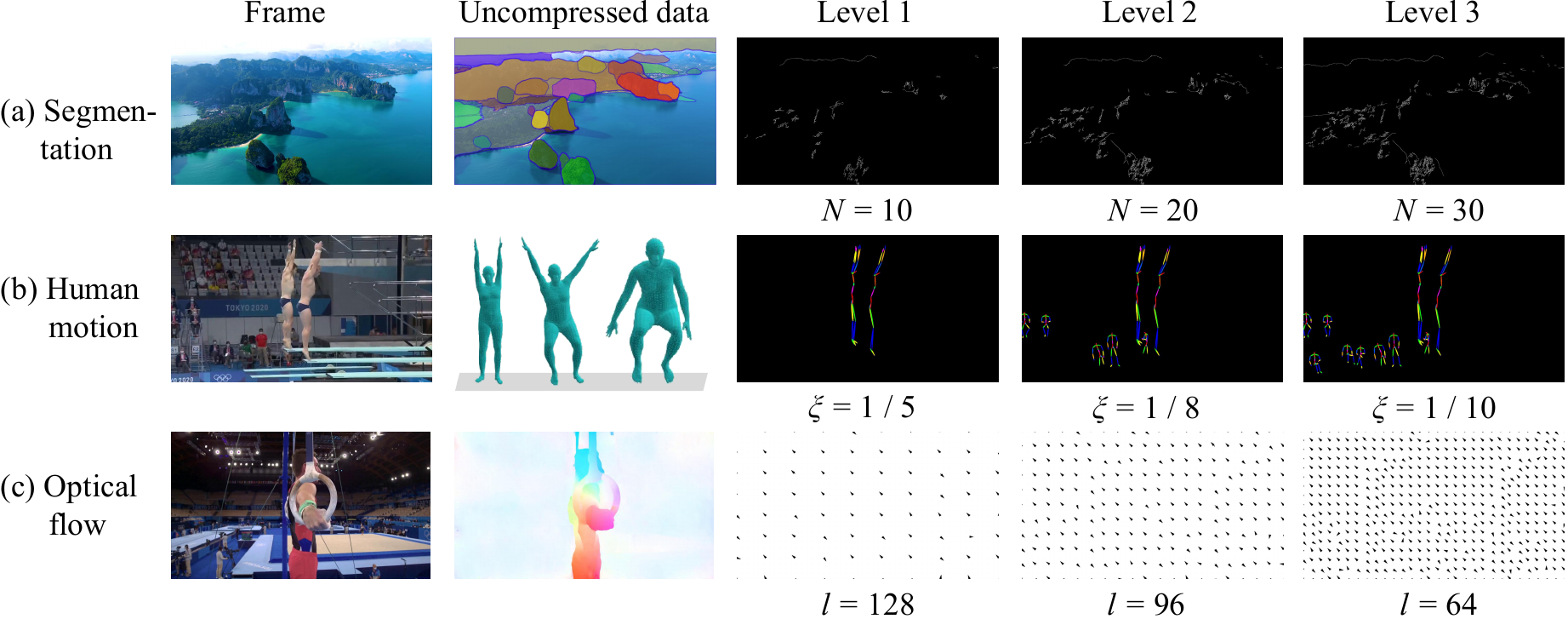}
    \vspace{-2mm}
    \caption{Visualization of optical flow, human motion, and segmentation representations alongside their compact forms at varying bitrate thresholds.}
    \label{fig:demo}
\end{figure*}

Our conditional diffusion model is built upon the pretrained FL2V diffusion model WAN2.1~\citep{wan2025}. Firstly, the condition signals, i.e. segmentation/human motion/optical flow are convert to dense visual modalities $V=\{V_{seg}, V_{motion}, V_{flow} \} \in R^{T * H * W * 3}$, as shown in Fig.~\ref{fig:demo}. Given the paried visual modalities, we first encode them into a latent space using a pretrained 3D causal VAE encoder $\epsilon$.
$$x_m=\epsilon(V_m), m \in \{flow, motion, seg\}$$

During training, a key challenge arises from modality entanglement: the model often over-relies on dominant conditions (e.g., segmentation) while neglecting others. To enforce balanced utilization, We apply \textbf{random dropout} for each condition with a dropout ratio of 0.3. For any condition subjected to dropout, its visual representation is replaced with a zero-valued (all-black) image sequence. 

Meanwhile, random dropout introduces an another problem: \textbf{role ambiguity}. 
Under extreme transmission conditions, certain conditioning signals may be lost or become incomplete, such as the absence of pose control. However, the model is unable to distinguish whether the missing condition results from accidentally absent conditions or from dropout-induced zeros.
To address this, we introduce an adaptive control strategy that dynamically assigns roles to different modalities. 
This idea is first introduced by OmniVDiff~\citep{xi2025omnivdiff} to distinguish whether the input is intended for joint generation or merely as conditioning information. Distinct from their objective, we introduce a modality embedding $e_m$ that differentiates between the roles of \textit{dropout} ($e_d$) and \textit{conditioning} ($e_c$), which can be directly added to the latent space.


 \begin{equation}
x_m = \bar{x}_m + \big[ \delta_m \, e_d + (1-\delta_m) \, e_c \big],
\end{equation}

where $\delta_m$ is a binary indicator taking values 0 or 1, specifying whether dropout is applied to $m$:  

$\delta_m = 1 \;\;\Rightarrow\;\;$ the dropout offset $e_d$ is used,  

$\delta_m = 0 \;\;\Rightarrow\;\;$ the conditioning offset $e_c$ is used.

Finally, we concatenate $x_m$ along the channel dimension with the latent noise, forming the input of the diffusion transformer. We freeze the majority of the diffusion model's parameters and perform efficient adaptation via Low-Rank Adaptation (LoRA)~\citep{hu2022lora}.

This strategy enables flexible and efficient control, allowing the model to to remain robust even when certain signals are unavailable.

\section{Experiments}
For training the multi-condition diffusion model, we employ a curated subset of 20,000 diverse video sequences sampled from Koala-36M~\citep{wang2024koala}. To evaluate the performance of our method and state-of-the-art (SOTA) methods on the video compression task, we conduct a series of experiments and empirical studies on public datasets including HEVC Class B, C~\citep{HEVC_class_B_and_C} as well as UVG~\citep{Mercat_2020_UVG} and MCL-JCV~\citep{Wang_2016_MCL_JCV}, following~\citep{wan2024m3cvccontrollablevideocompression} and T-GVC~\citep{Wang_2025_T_GVC}. Furthermore, to address the limited scene coverage in open-source benchmarks, we conduct rigorous cross-dataset evaluation on our curated internet-sourced video corpus. Extensive experiments validating our method's generalization capability across diverse real-world scenarios are provided in the supplementary material.

\subsection{Experimental Settings}




\noindent \textbf{Evaluation Metrics.}
We use two commonly used metrics for evaluation: \textbf{Fréchet Video Distance} (\textbf{FVD}) and \textbf{Learned Perceptual Image Patch Similarity} (\textbf{LPIPS}), as they better align with human perception compared to traditional measures such as PSNR and SSIM. Lower FVD and LPIPS at the same bitrate represent better performance.



\noindent \textbf{Baselines.}
To demonstrate the effectiveness of the framework, we compare the proposed framework with both traditional video compression standards, H.264, H.265, H.266, and SOTA video compression methods:  (1) traditional video compression methods (H.264~\citep{Wiegand_2003_H264}, H.265~\citep{Sullivan_2012_H265}, H.266~\citep{Bross_2021_H266}); (2) neural video compression methods (DCVC-DC~\citep{li2023neural}, DCVC-FM~\citep{li2024neural}, DCVC-RT~\citep{jia2025towards}); (3) diffusion-based video compression methods (T-GVC).

\begin{table}[H]
    \centering
    \begin{tabular}{cccc}
         \toprule
         Level & Segmentation & Human Motion & Optical Flow \\
         \midrule
         Level 0 & N/A & N/A & N/A \\
         Level 1 & $N = 10$ & $\xi = 1 / 5$ & $l = 128$ \\
         Level 2 & $N = 20$ & $\xi = 1 / 8$ & $l = 96$ \\
         Level 3 & $N = 30$ & $\xi = 1 / 10$ & $l = 64$ \\
         \bottomrule
    \end{tabular}
    \caption{Different compression settings.}
    \label{tab:compression}
\end{table}

\noindent \textbf{Compression Settings.}
To demonstrate the effectiveness of our framework comprehensively, we design multiple compression settings, as shown in Table~\ref{tab:compression}. Specifically, for Level 0, we only use the first and the last frame of the clip, with no additional conditions provided.  During training, we randomly select one configuration from multiple compression settings to process the training data, enabling a single model to simultaneously handle inputs with different settings.

\begin{table*}[ht]
    \centering
    \resizebox{0.99\textwidth}{!}{

    \begin{tabular}{lcccccccc}
         \toprule
         \multirow{2}{*}{Method} & \multicolumn{2}{c}{HEVC Class B} & \multicolumn{2}{c}{HEVC Class C} & \multicolumn{2}{c}{UVG} & \multicolumn{2}{c}{MCL-JCV}\\
         \cmidrule{2-9} & FVD (↓) & LPIPS (↓) & FVD (↓) & LPIPS (↓) & FVD (↓) & LPIPS (↓) & FVD (↓) & LPIPS (↓)\\ 
         \midrule
         H.264~\citep{Wiegand_2003_H264} & 2738 & 0.8283 & 3022 & 0.7724 & 4252 & 0.8274 & 4844 & 0.7139 \\
         H.265~\cite{Sullivan_2012_H265} & 1327 & 0.4941 & 1452 & 0.4833 & 1688 & 0.4052 & 1030 & 0.3667 \\
         H.266~\citep{Bross_2021_H266} & 1022 & 0.4038 & 1273 & 0.4543 & 1052 & 0.2834 & 973 & 0.3217 \\
         DCVC-DC~\citep{li2023neural} & 892 & 0.3821 & 1135 & 0.4327 & 992 & 0.2742 & 887 & 0.3152 \\
         DCVC-FM~\citep{li2024neural} & 837 & 0.3672 & 1079 & 0.4186 & 968 & 0.2689 & 849 & 0.3103 \\
         DCVC-RT~\citep{jia2025towards} & 783 & 0.3543 & 1029 & 0.4058 & 947 & 0.2635 & 811 & 0.3068 \\
         T-GVC~\citep{Wang_2025_T_GVC} & - & 0.3512 & - & 0.3642 & - & 0.2212 & - & 0.3145\\
         \textbf{Ours} & \textbf{597} & \textbf{0.2813} & \textbf{402} & \textbf{0.2625} & \textbf{571} & \textbf{0.2208} & \textbf{515} & \textbf{0.2926} \\
         \bottomrule
    \end{tabular}
    }
    \caption{The overall performance of different methods on the test sets (BPP = 0.0066). Here, we report the FVD and LPIPS among actual videos and predictions generated by different methods. The best results are highlighted in \textbf{bold}.\label{tab:sota}}
\end{table*}

\subsubsection{Hyperparameter Settings.}
For a fair comparison, all methods are implemented with PyTorch 2.5 in Python 3.9.13 and learned with Adam optimizer~\citep{Kingma14_Adam}.
We conduct our experiments on a single Linux server with 2 Intel(R) Xeon(R) CPU Platinum 8558 @2.10 GHz, 2 TB RAM, and 8 NVIDIA H100 (80 GB of graphic memory each). We fine-tune the pre-trained Wan2.1-14B for 3 epochs with a learning rate of $2 \times 10^{-5}$ and a batch size of 8. We tuned the parameters of all methods over the validation set. We set rank (\emph{r}) and alpha ($\alpha$) to 16 for LoRA.

\subsection{Overall Performance}
\begin{figure*}[ht]
    \centering
    \includegraphics[width=0.99\textwidth]{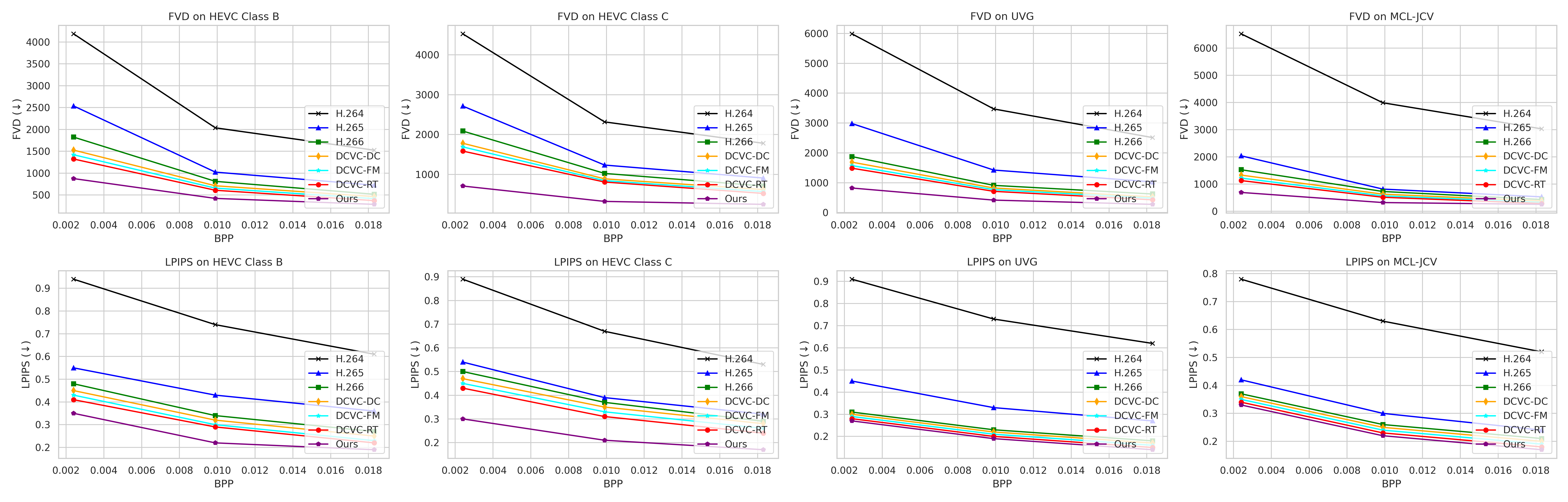}
    \vspace{-2mm}
    \caption{R-D (perception) performance with state-of-the-art models. \textbf{Upper}: FVD (↓) on HEVC-B/C, UVG and MCL-JCV datasets; \textbf{Lower}: LPIPS (↓) on HEVC-B/C, UVG and MCL-JCV datasets.}
    \label{fig:sota}
\end{figure*}

Table~\ref{tab:sota} summarizes the performance of all models on Level 1 (BPP = 0.0066). Please note that since T-GVC has not released its codebase, we directly use the performance reported in the paper~\citep{Wang_2025_T_GVC}. Fig.~\ref{fig:sota} summarizes the overall performance of all models on other levels (Level 0: BPP = 0.0024, Level 2: BPP = 0.0099, Level 3: BPP = 0.0183). 
Here, we make the following observations. 


First, our diffusion-based video compression framework (conditioned on human motion, canny edges, and optical flow) outperforms traditional codecs (H.264/H.265/H.266) and neural compression baselines (e.g., DCVC-RT) in all bitrate settings. Quantitative metrics (FVD and LPIPS) show improvements of 15–30\%, particularly in extreme low bitrate settings, confirming that the generated videos better preserve perceptual quality, avoiding blocking artifacts (common in traditional codecs) and over-smoothed textures (typical of neural methods).

Second, while higher compression ratios lead to lower objective scores, the generative nature of diffusion ensures graceful degradation in perceptual quality. Even at level 1 compression where bpp is less than 0.007, key motion and semantics remain recognizable (see visualized results), making the method viable for bandwidth-constrained applications (e.g., mobile streaming, surveillance).

Last, traditional codecs like H.264 fail to produce results at extreme low bitrates, exhibiting severe blocking artifacts and temporal inconsistency. In contrast, our method maintains coherent motion and sharp edges even at Level 1 (BPP = 0.0066), demonstrating superior robustness across bitrates.


\subsection{Ablation Study}
\begin{figure}[t]
    \centering
    \includegraphics[width=0.49\textwidth]{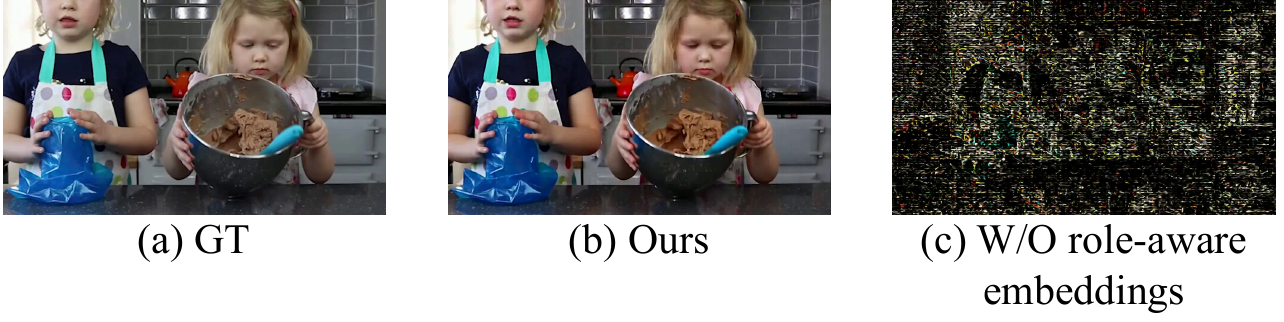}
    \vspace{-4mm}
    \caption{Ablation study on role-aware embeddings. }
    \label{fig:ablation}
\end{figure}

To validate the necessity of our core innovations, stochastic modality dropout and role-aware embeddings, we conduct ablation experiments with a baseline model, which is trained without both modality dropout and role-aware embeddings. We perform the study by systematically removing individual modalities—segmentation sequences (Seg), human motion (Motion), and optical flow (Flow). It worth noting that, when modality is ablated, its visual representation is replaced with a zero-valued (all-black) image sequence. Meanwhile, for our model, the modality embedding $e_m$ is assigned to the dropout role $e_d$, which then will be added in the latent space. We have two observations:
Firstly, as shown in Fig.~\ref{fig:ablation}(c),  when one modality is missing, baseline model causes catastrophic failure and yields noise-like outputs.
Secondly, as shown in Table~\ref{tab:ablation}, our model achieves best results across both datasets when all modalities are valid, while removing any single condition degrades the FVD.

The ablation experiments demonstrate that our multi-condition training paradigm not only prevents the model from over-relying on any single cue, but also enables the composition of arbitrary condition subsets at the decoder to faithfully emulate degradation patterns encountered in real-world scenarios.

	

\begin{table}[t]
	\caption{The ablation study of condition settings on the UVG and MCL-JCV datasets at Level 1 (BPP = 0.0066). The best results are highlighted in \textbf{bold}.\label{tab:ablation}}
	\centering
	\resizebox{0.47\textwidth}{!}{
	\begin{tabular}{ccccc}
        \toprule
        \multirow{2}{*}{Condition} & \multicolumn{2}{c}{UVG} & \multicolumn{2}{c}{MCL-JCV}\\
        \cmidrule(lr){2-5} & FVD (↓) & LPIPS (↓) & FVD (↓) & LPIPS (↓)\\
        \midrule
        w/o Seg    & 623          & 0.2481          & 577          & 0.3102 \\
        w/o Motion & 605          & 0.2353          & 618          & 0.3369 \\
        w/o Flow   & 589          & 0.2308          & 542          & 0.3015 \\
        full model & \textbf{571} & \textbf{0.2208} & \textbf{515} & \textbf{0.2926} \\
	\bottomrule
	\end{tabular}
        }
\end{table}





\subsection{Visualized Results}
\label{sec: visualized_results}
\begin{figure}[t]
    \centering
    \includegraphics[width=0.49\textwidth]{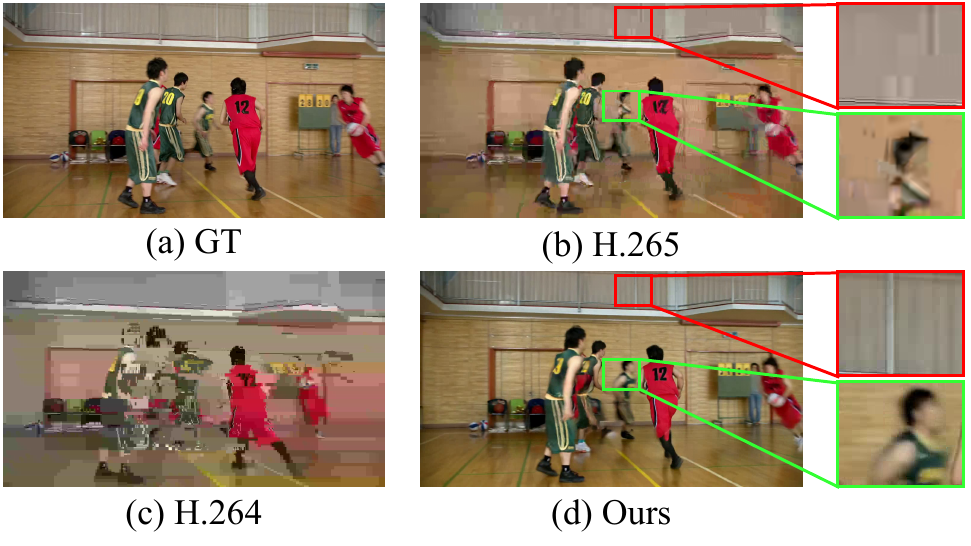}
    \vspace{-6mm}
    \caption{Visual comparison with SOTA codecs on the HEVC Class B set. We present the frame between original video, H.264/H.265, and our model. All predicted results are controlled to have a bpp of 0.0099 (Best viewed zoomed in and in color).}
    \label{fig:vis}
\end{figure}
In this subsection, we present some visualized results to demonstrate the performance of our model and comparative methods intuitively.

Fig.~\ref{fig:vis} provides a qualitative comparison of video compression performance across different methods, specifically between Ours, H.264/H.265, and the Ground Truth. The selected sample illustrates that our method preserves semantic and structural details even at ultra low bitrates, whereas H.264 and H.265 exhibit significant compression artifacts including noticeable blurring effects.

\section{Conclusion}
This paper presents a video compression framework leveraging conditional diffusion models to achieve human-perception-aligned reconstruction at ultra-low bitrates. Despite these advances, limitations remain: The current decoding speed falls short of real-time requirements due to computational intensity in diffusion-based generation. To address this, we propose adopting \textit{familiar model}—deploying homologous diffusion models of varying capacities—enabling automatic model switching based on decoder-side computational resources, inspired by AI Flow paradigms~\citep{an2025aiflowperspectivesscenarios}. Additionally, content-agnostic bitrate control via manual parameter tuning (e.g., $l$, $N$, $\xi$) limits adaptability. Future work will integrate content-understanding modules to dynamically optimize compression thresholds based on scene complexity and motion characteristics.

\bibliographystyle{aaai} 
\bibliography{aaai}

\end{document}